\documentclass[letterpaper]{article} 
\usepackage{aaai24}  
\usepackage{times}  
\usepackage{helvet}  
\usepackage{courier}  
\usepackage[hyphens]{url}  
\usepackage{graphicx} 
\usepackage{natbib}  
\usepackage{caption} 
\usepackage{multirow}
\usepackage{algorithm}
\usepackage{algorithmic}

\usepackage{newfloat}
\usepackage{listings}
\DeclareCaptionStyle{ruled}{labelfont=normalfont,labelsep=colon,strut=off} 
\floatstyle{ruled}
\newfloat{listing}{tb}{lst}{}
\floatname{listing}{Listing}
\title{My Art My Choice: Adversarial Protection Against Unruly AI}
\author{
    Anthony Rhodes\textsuperscript{\rm 1}, Ram Bhagat\textsuperscript{\rm 2}, Umur Aybars \c{C}ift\c{c}i\textsuperscript{\rm 2}, \.{I}lke Demir\textsuperscript{\rm 1} 
}
\affiliations{
    \textsuperscript{\rm 1}Intel Labs, Santa Clara, CA\\
    \textsuperscript{\rm 2}Binghamton University, Binghamton, NY\\
    \{anthony.rhodes, ilke.demir\}@intel.com, \{rbhagat2, uciftci\}@binghamton.edu

}

\begin{document}
\twocolumn[{%
\renewcommand\twocolumn[1][]{#1}%
\maketitle
\vspace{-0.5in}
\begin{center}
    \includegraphics[width=1\textwidth]{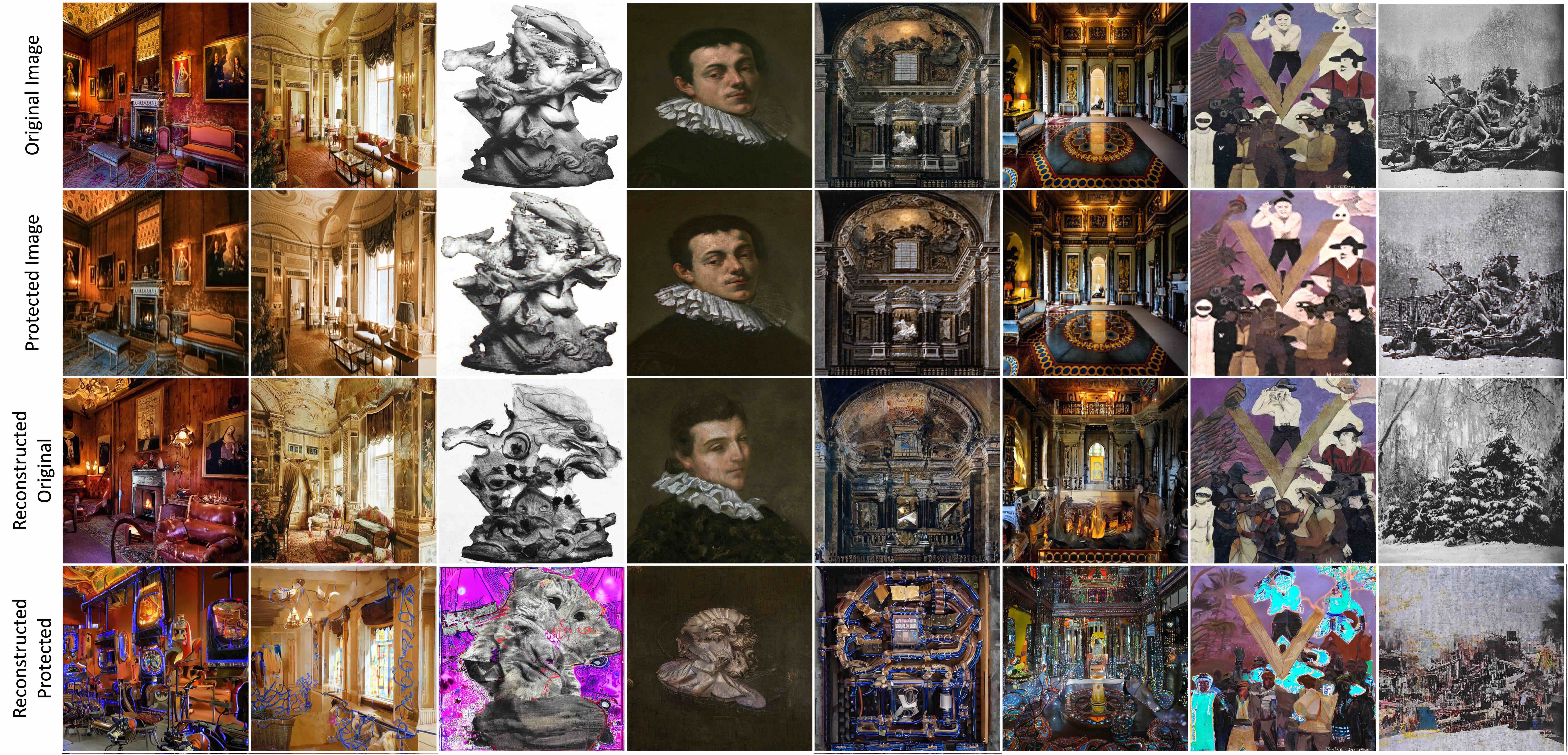}\\
    \textbf{Figure 1. }{Our approach enables artists to protect their content (first row) by learning to create perturbed versions (second row). Diffusion models exploit the original artwork (third row), however, protected images break these models (last row).}
    \label{fig:teaser}
\end{center}%
\stepcounter{figure}
}]

\begin{abstract}
Generative AI is on the rise, enabling everyone to produce realistic content via publicly available interfaces. Especially for guided image generation, diffusion models are changing the creator economy by producing high quality low cost content. In parallel, artists are rising against unruly AI, since their artwork are leveraged, distributed, and dissimulated by large generative models. Our approach, My Art My Choice (MAMC), aims to empower content owners by protecting their copyrighted materials from being utilized by diffusion models in an adversarial fashion. MAMC learns to generate adversarially perturbed ``protected'' versions of images which can in turn ``break'' diffusion models. The perturbation amount is decided by the artist to balance distortion vs. protection of the content. MAMC is designed with a simple UNet-based generator, attacking black box diffusion models, combining several losses to create adversarial twins of the original artwork. We experiment on three datasets for various image-to-image tasks, with different user control values. Both protected image and diffusion output results are evaluated in visual, noise, structure, pixel, and generative spaces to validate our claims. We believe that MAMC is a crucial step for preserving ownership information for AI generated content in a flawless, based-on-need, and human-centric way.
\end{abstract}

\section{Introduction}
Generative modeling has been introduced over half a century ago, with applications in mathematics~\cite{snyder1992generative}, shapes~\cite{ebert2000procedural}, botany~\cite{aono1984botanical}, architecture~\cite{demir2016proceduralization}, and many other domains. Recently, deep counterparts of generative models, or generative AI, are proliferating as a hyper-realistic way of creating visual content, proportionately to the democratization of generative AI services, such as Stable Diffusion~\cite{sd}. From a cheap content production perspective, this has been the moment all creatives were waiting for. However, these large models are often trained on large amounts of internet-scrapped visual data, disregarding any ownership or copyright of the materials. As these models learn from the data, they are able to mimic specific content, style, or structure of training samples \--- replicating art. Consequently, artists and creators are resisting against unruly use of AI~\cite{getty, sdsue, artiststrike}, since (1) generative AI creates derivatives of their art without liabilities, (2) diffusion models are trained on their data without permission, and (3) there is no compensation mechanism for outputs, replicating their art.
\begin{figure*}[ht]
\centering
\includegraphics[width=1\linewidth]{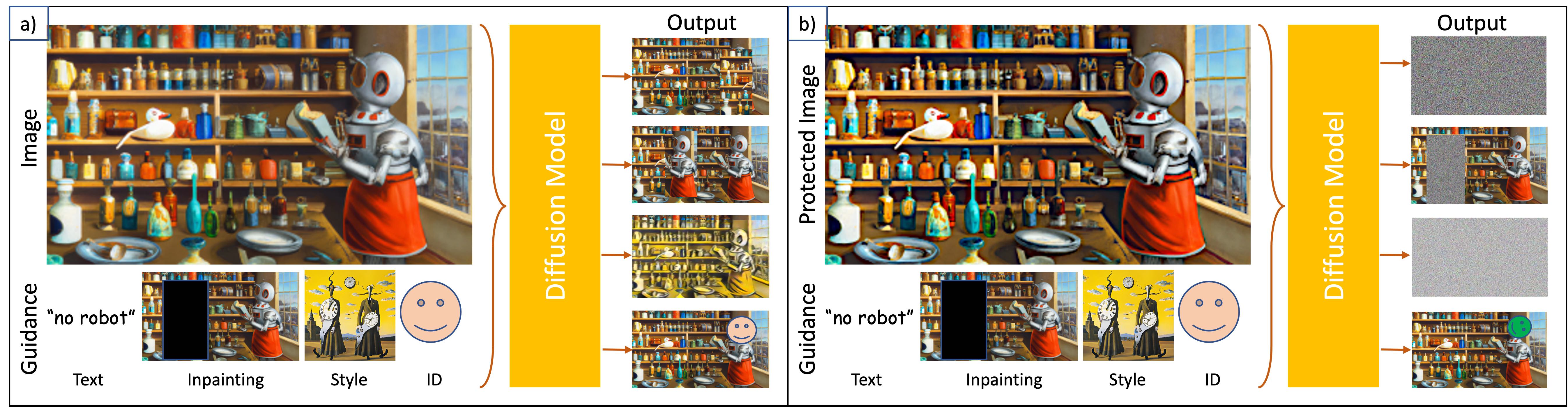} 
\caption{MAMC aims to learn how to confuse diffusion models for degrading their  output realism for various tasks.}
\label{fig:motivation}
\end{figure*}

As regulation and policy space is not mature enough to protect creative rights, we propose an interim AI tool to enable artists to seal their material with adversarial protection against generative AI. Colloquial to its name,``My Art My Choice'' provides artists the choice to protect their content and the strength of protection, from being used in generative AI applications. When these imperceptibly different versions of original images are fed to diffusion models for various tasks, we aim to have a garbage output, similar to Fig.~\ref{fig:motivation}. Unlike previous work~\cite{glaze}, (1) we do not utilize a driving image, (2) we let the artist set the amount of distortion, and (3) we define a multi-objective training regime for better quality and more atrophy.

We model this problem as an adversarial attack to black box diffusion models where only the input/output pairs are known in various tasks. We optimize for the perceptual resemblance of the input and protected images, and structural and generative atrophy of the diffusion output image. By design, MAMC aims the artist to be in control, thus, we craft and expose a balance variable for the artist to tune between the fidelity and robustness of the protected image. Our main contributions include,
\begin{itemize}
\item a model that learns to adversarially protect any given image against diffusion models, 
\item a human-centric AI system to balance protection and perturbation of the content, and
\item experimental validation of the approach in various generative AI tasks and datasets.
\end{itemize}

We demonstrate our results on three datasets, qualitatively and quantitatively comparing (1) input and protected images, and (2) diffusion model outputs of input and protected images, as demonstrated in Fig.1. Different losses, hyperparameters, and generative AI tasks are also experimented with in our results section. Finally, effects of the balance variable are analyzed to document user influence.

\section{Related Work}
\subsection{Controlled Content Generation}
Image~\cite{glassner2014principles}, texture~\cite{efros1999texture}, shape~\cite{kalogerakis2012probabilistic}, and appearance~\cite{zhou2016view} synthesis has been explored for decades by generative modeling. Introduction of Generative Adversarial Networks (GAN) in 2014~\cite{GAN} ignited a surge of photorealistic content generation approaches, applications, and services; followed by Diffusion Models (DM)~\cite{diffusion}, surpassing GANs in terms of quality of and control over generated content~\cite{survey}.

DMs are first proposed for text-to-image generation, an image is sampled from the learned distribution given a text prompt. Especially personalized text-to-image generation approaches learn generating specific identity, style, or context with a few input images via fine-tuning~\cite{dreambooth, Mokady_2023_CVPR} or by tokenization~\cite{instantbooth}, as image guidance emerges for DMs. Recently, DMs are used for crafting stories~\cite{talecrafter}, deepfakes~\cite{hyperdreambooth}, multi-person images~\cite{fastcomposer}, pasting objects into scenes~\cite{paintbyex,zhang2023paste}, image editing~\cite{instructpix2pix,zeropix2pix,sine,controlnet}, object editing~\cite{pairdiff}, video synthesis~\cite{align}, 3D avatars~\cite{rodin}, and many other applications; all of which require additional image input for guidance. MAMC aims to protect this guiding image. For example, if an image of the character in the crafted story is guiding generation, MAMC aims that the output scenes do not contain the character. If several face images are given to personalize a diffusion model for deepfakes, MAMC aims that the diffusion output does not look like a face. If a scene is inpainted with a given object, MAMC aims that the scene does not contain that object. If the diffusion model is controlled by a style image, MAMC aims that neither style nor structure is preserved. In summary, MAMC aims to ``break'' the diffusion model by pushing the output far away from an ideally expected output. 

\subsection{Adversarial Generation}
There has been a considerable amount on literature for using image manipulation and generation techniques in adversarial settings~\cite{attacksurvey}, including those against face recognition~\cite{clip2protect, mfmc, lowkey}, breaking deepfake detectors~\cite{sophie}, and preemptively confusing models by data augmentation~\cite{dataaugmentation}. These approaches utilize generative models to attack other deep learning models, however, the construction of adversarial generation may be similar to adversarial protection: The loss is guided by the black box model's response to adversarially perturbed inputs. To clarify this distinction, MAMC uses a generative model to attack active/synthesis models, whereas others listed use a generative model to attack passive/analysis models. This adds another layer of complexity to define our threat model, i.e., what does it mean to `break a diffusion model' vs. `break a face recognizer', as it will be explained later.

\subsection{Adversarial Protection}
Finally, we would like to categorize approaches that use a generative model to attack a generative model, under adversarial protection. Most of the aforementioned DMs are trained on large internet scraped visual datasets such as LiASON~\cite{liason} without any ownership or copyright monitoring. As a result, DMs can replicate the content~\cite{somepalli2023diffusion}, style~\cite{zhang2023inversion}, and structure~\cite{controlnet} of samples; which is violating artists' rights over their own materials. Emerging research battles with this problem by machine unlearning~\cite{unlearning} to delete samples from DMs manifold, by confusing the model to converge towards a different style target~\cite{glaze} to prohibit style mimicry, by focusing on disabling specific DMs~\cite{antidreambooth} such as~\cite{dreambooth}, by injecting noise into input images~\cite{photoguard} to disable text-based image editing, and by Compartmentalized Diffusion Models~\cite{golatkar2023} to selective forgetting in a continual learning setting. MAMC follows this route to provide adversarial protection, (1) by learning to generate imperceptible adversarial twins, (2) using a combination of losses robust against several tasks, and (3) with external controllable balancing between distortion and protection of an image. Unlike previous work mentioned above, MAMC is not limited to specific tasks, models, or domains.

\section{My Art My Choice}
Our system learns to generate a perturbed version of any given image to fool diffusion models when the perturbed image is used as source or guidance. Overview of the system is depicted in Fig.~\ref{fig:system}.
\subsection{Problem Formulation}
We design MAMC to be a generalized adversarial protector, thus we formalize our assumptions as follows:
\begin{itemize}
    \item Content owners want to share their images online, however, they do not want these images to be trained on, replicated, or influenced by generative models.
    \item Content owners introduce imperceptible adversarial changes to cloak their artwork before its release.
    \item Artists want to have control over how much their images are distorted due to protection.
    \item The adversary has access to these images and aims to use them to train, fine-tune, or guide diffusion models; so that the content, style, identity, or structure of these images can be replicated.
    \item The adversary is not aware of our protection mechanism. 
    \item Both artists and adversary can query a pre-trained diffusion model and have access to reasonable compute resources. Note that neither has access to the model weights or resources to train a diffusion model from scratch.
\end{itemize}
Under these assumptions, given an image $I$, MAMC learns to generate $G(I)=I+\delta=I'$ where $\delta$ is the learned perturbation to attack a black-box diffusion model $M$. This attack should create an image as dissimilar to the expectations as possible, meaning that $M(I)$ and $M(I')$ should be maximally dissimilar. Thus, adversarial protection optimization to learn this perturbation becomes
\begin{equation}
    \max_{\delta_I} ||M(I+\delta_I)-M(I)||, \ \ s.t. |\delta_I|<\phi_I+\epsilon
\end{equation}
where $\phi$ is the balance factor and $\epsilon$ represents a small neighborhood. Intuitively, we strive to push the diffusion model output as far as possible from the actual output. This enables generalization across tasks and models. Previous work~\cite{glaze} focuses only on the specific case of style transfer, where the expected output is pushed towards a dissimilar target style. The similarity and representativeness of this target style plays a significant role for their minimization to converge on a meaningful dissimilar style result. Our approach is free of any such selection.

\subsection{Architecture}
\begin{figure}[h]
\centering
\includegraphics[width=1\columnwidth]{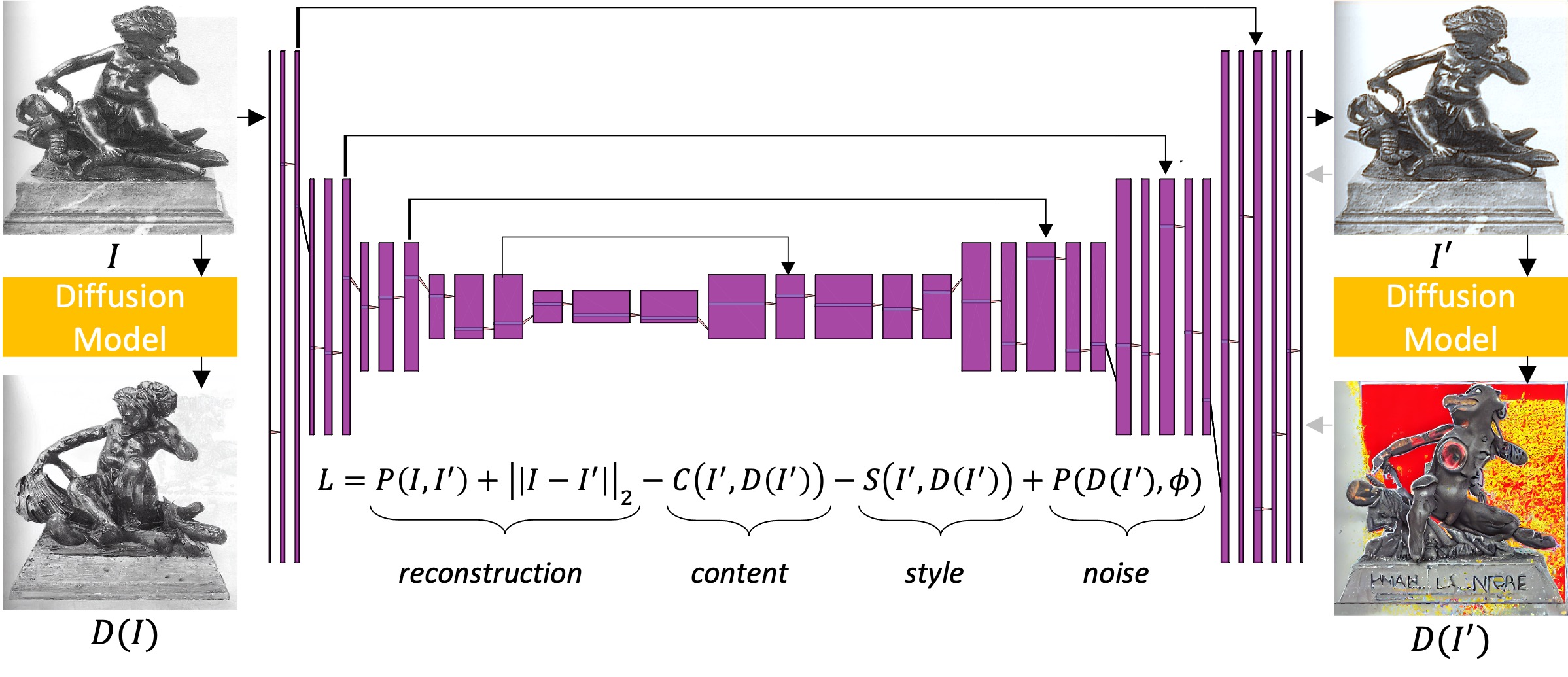} 
\caption{Our input and output samples, generator architecture, and loss formulation is simplified in this overview.}
\label{fig:system}
\end{figure}

We employ a simple UNet architecture~\cite{unet} to learn this generation process, consisting of blocks with two convolutional layers followed by up/downsampling, with concatenations between every encoder/decoder block (Fig.~\ref{fig:system}). We use a standard pre-trained diffusion model~\cite{stablediffusion} in frozen state to infer input output relations.
\subsection{Training Objective}
We formulate a multi-objective function that combine additional losses to satisfy the initial assumptions.
\subsubsection{Reconstruction Loss.} As artists expect minimal changes, we introduce a reconstruction term $L_R$, keeping input and protected images perceptually similar. We use LPIPS~\cite{lpips} for perceptual similarity ($\mathcal{P}$). We also add a pixel-wise $\ell_2$ norm to prevent color shifts.
\begin{equation}
    L_R = \alpha_{R1}\mathcal{P}(I, I') + \alpha_{R2}||I-I'||^2_2
\end{equation}
\subsubsection{Content Loss.} For inpainting and personalization use of diffusion models, content of the input should not be preserved. Thus, we introduce a content loss where diffusion output is not perceptually similar to the protected image.
\begin{equation}
    L_C = - \alpha_{C}\mathcal{P}(I', M(I'))
\end{equation}
\subsubsection{Style Loss.} In order to prohibit style transfer and reconstruction applications of diffusion models, we introduce a style loss as the distance between Gram Matrices $\Omega$ of the protected image and its diffusion output, over activations $j$.
\begin{equation}
    L_S = - \alpha_{S}\frac{1}{|j|}\sum_j||\Omega_j(I') - \Omega_j(M(I'))||
\end{equation}
\subsubsection{Noise Loss.} Finally, to really confuse the diffusion model, we introduce a noise loss to put diffusion output of the protected image towards Gaussian noise, indicated by $\mathcal{N}$.
\begin{equation}
    L_N = \alpha_{N1}\mathcal{P}(M(I'), \mathcal{N})
\end{equation}
\subsubsection{Multi-objective Loss.} To sum up, our overall loss function is constructed as follows. We set loss weights $\alpha_*$ experimentally, following our ablation studies in Sec.~\ref{sec:abla}.
\begin{equation}
    L = \alpha_{R}L_R - \alpha_{C}L_C - \alpha_{S}L_S + \alpha_{N}L_N
\end{equation}
\subsection{Balance Factor}
Unlike traditional generative models, deep generative models are not \textit{WYSIWYG}, thus, their outputs tend to not overlap with expectations. As MAMC is also a generative model, we aim to provide some control to the content owners, especially if we are altering their material. To that front, we experiment with predefined sets of $\alpha_*$ hyper-parameters and prepare MAMC versions with different strengths. These models are then exposed to the user based on demand for enabling them to balance distortion vs. protection of their images. We investigate these models in Sec.~\ref{sec:user}.

\section{Results}
We present evaluations and experiments of My Art My Choice on four datasets for a comprehensive understanding across domains: Wiki Art~\cite{wikiart} with 1K and 5K subsets, Historic Art~\cite{historicart} with 1K and 5K subsets, Art 201~\cite{} with 200 images, and FaceForensics++~\cite{ff} with 100 images. We select these as representative datasets, covering diverse content, style, artist, and domain fronts. We use 70\%-30\% train-test split for all of them and train the network for a small number of epochs as discussed in Sec.~\ref{sec:curve} with other implementation details. We compare My Art My Choice to some image cloaking methods in Appendix~\ref{app:A}.

\subsection{Use Cases}

\subsubsection{Protecting Artists from Style Infringement.} 
\begin{figure}[h]
\centering
\includegraphics[width=1\linewidth]{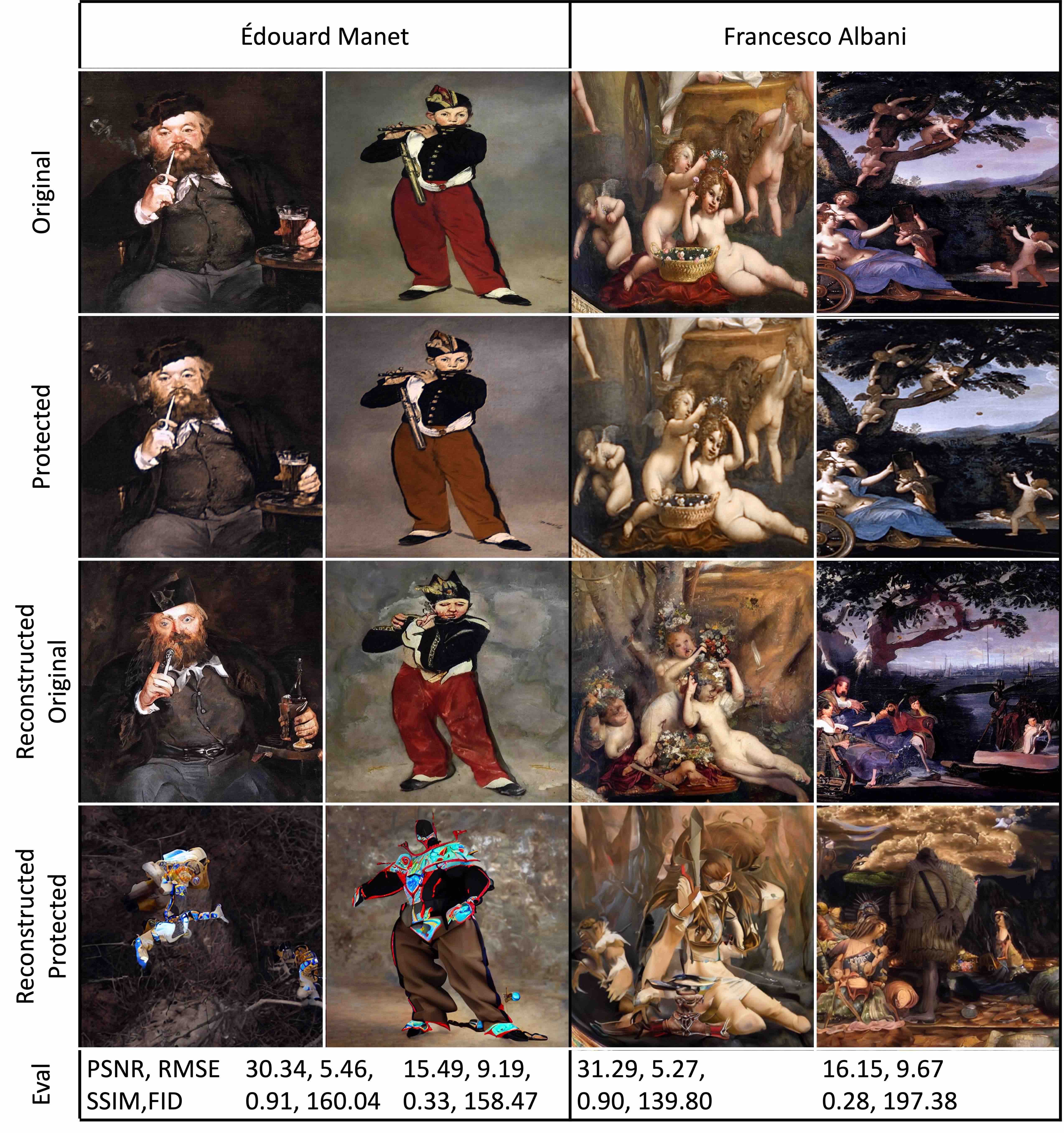} 
\caption{Diffusion models fail to replicate artists' style from adversarially protected images by MAMC.}
\label{fig:single}
\end{figure}

As creative industry is transforming, artists have been protesting against the use of generative AI. Specifically style transfer applications are claimed to steal artists' identity, as their style equates to their art. We evaluate MAMC on small single artist datasets to air its novelty. Fig.~\ref{fig:single} samples the works of Edouard Manet and Francesco Albani, shows easy replication of their style with diffusion models, and MAMC protected versions to disable that replication, along with the evaluation scores on the whole dataset. 

\subsubsection{Protecting Celebrities from Deepfakes.} 
\begin{figure}[h!]
\centering
\includegraphics[width=1\linewidth]{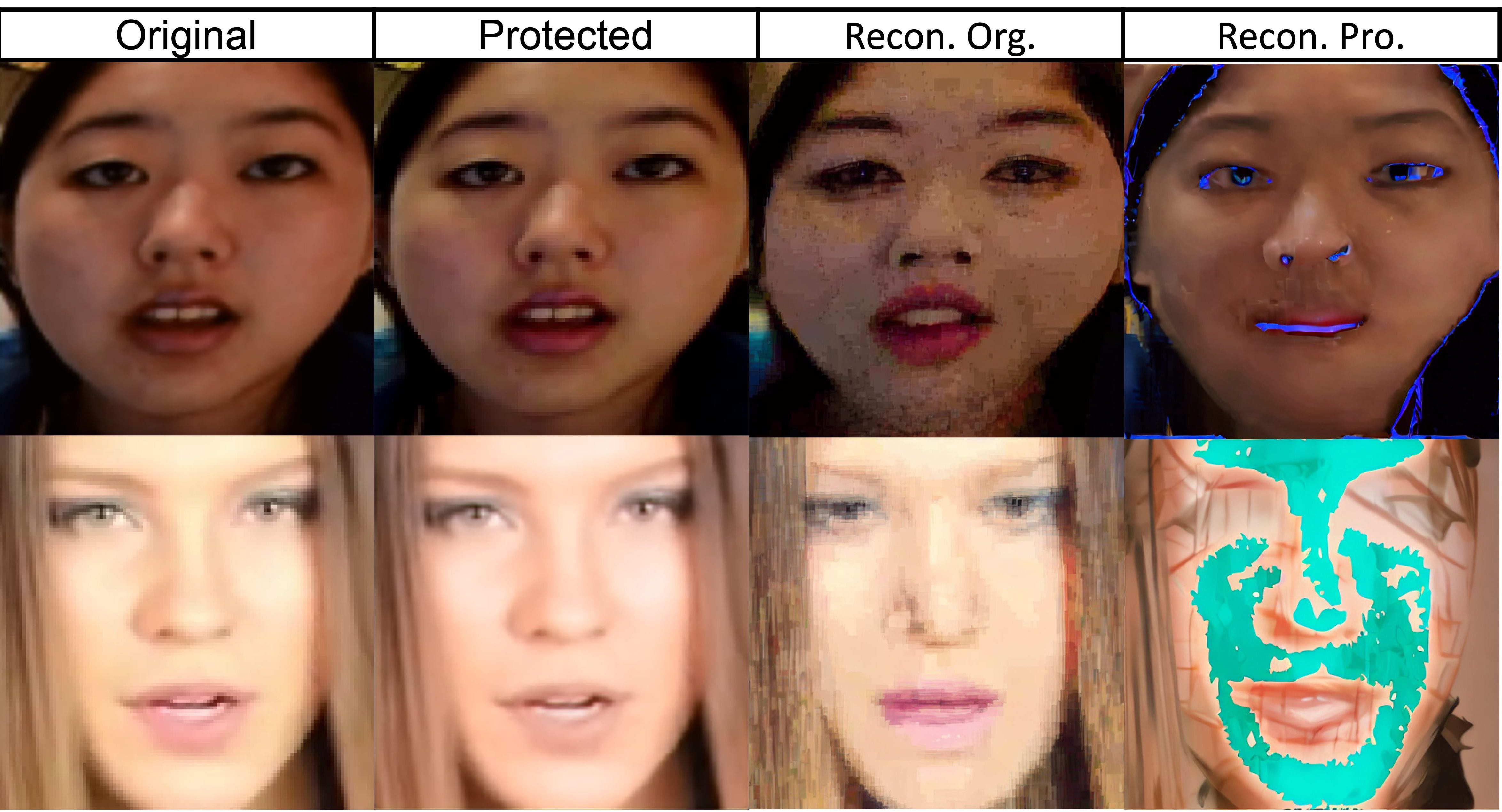} 
\caption{Diffusion models fail to replicate faces from adversarially protected images by MAMC.}
\label{fig:ff}
\end{figure}

Another popular use case of diffusion models is personalization, which means fine-tuning the model on a specific face to create look-alikes. We test MAMC on a face dataset to verify that it can also be used proactively against potential deepfakes. Fig.~\ref{fig:ff} depicts sample faces, their reconstructed, protected, and failed-to-reconstruct versions after altered with MAMC. The dataset evaluation scores between input and protected images, and their diffusion outputs are listed in Tab.~\ref{tab:all}. We leave the last use case as completing scenes and objects to Appendix~\ref{app:B}.

\begin{figure*}[h]
\centering
\includegraphics[width=1\linewidth]{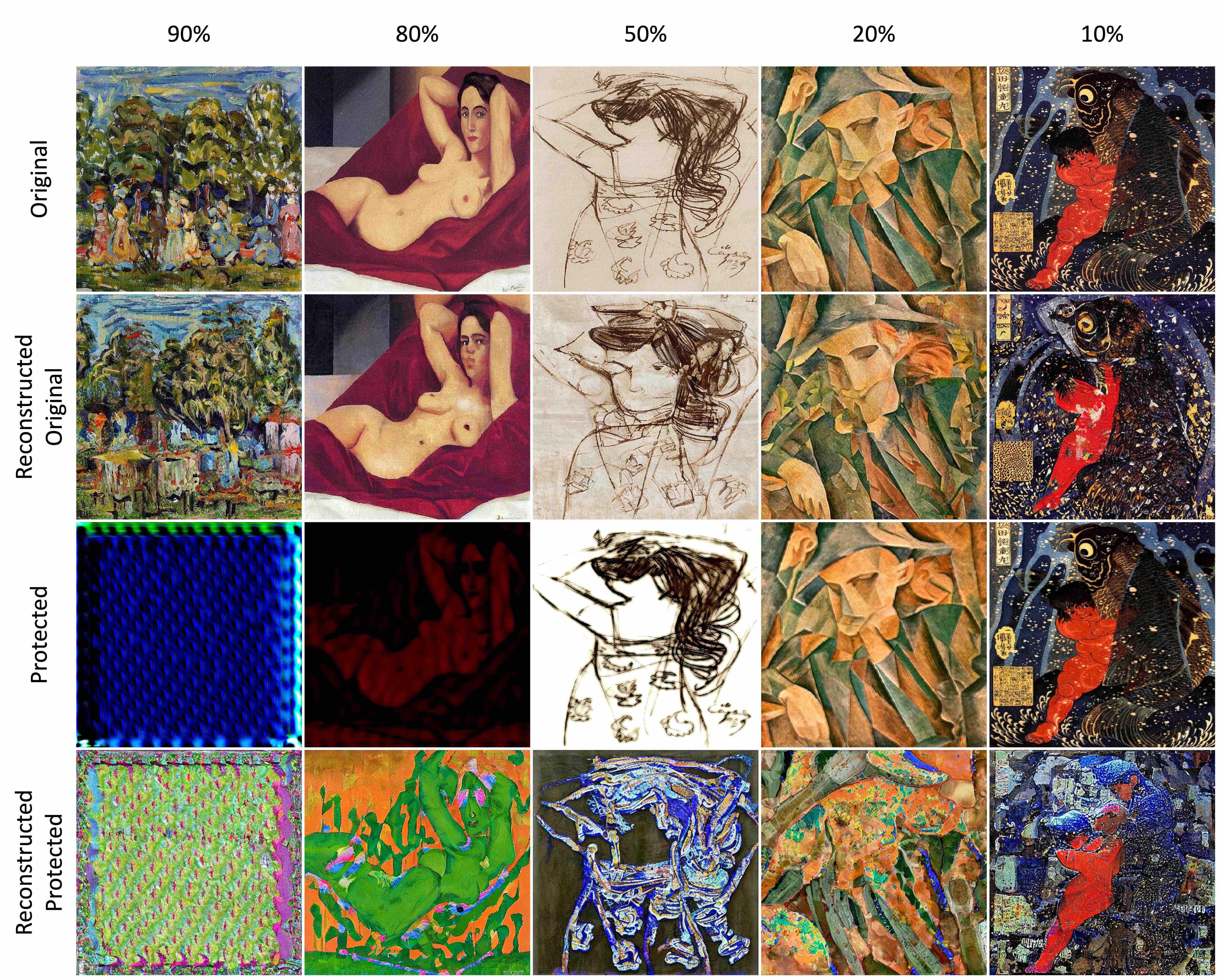} 
\caption{Analyzing the impact of the user exposed balance variable in different levels. The higher the percentage is the higher the protection and distortion are.}
\label{fig:user}
\end{figure*}

\subsection{User Control} \label{sec:user}
As mentioned, we would like the artist to have the freedom over how much preservation and protection is applied by MAMC. In Fig.~\ref{fig:user}, we demonstrate five levels of user controlled balance factor with varying levels of perturbations upon the artwork. As expected, when 90\% protection and distortion is desired, protected image does not look like the input image. Increasing it to 80\%, uncovers some input image features. At 50\%, we can see the image preserved with the protection and the protection yielding a very different diffusion output. As it increases, more details are preserved with still significantly different diffusion outputs.

\begin{figure*}[h]
\centering
\includegraphics[width=1\linewidth]{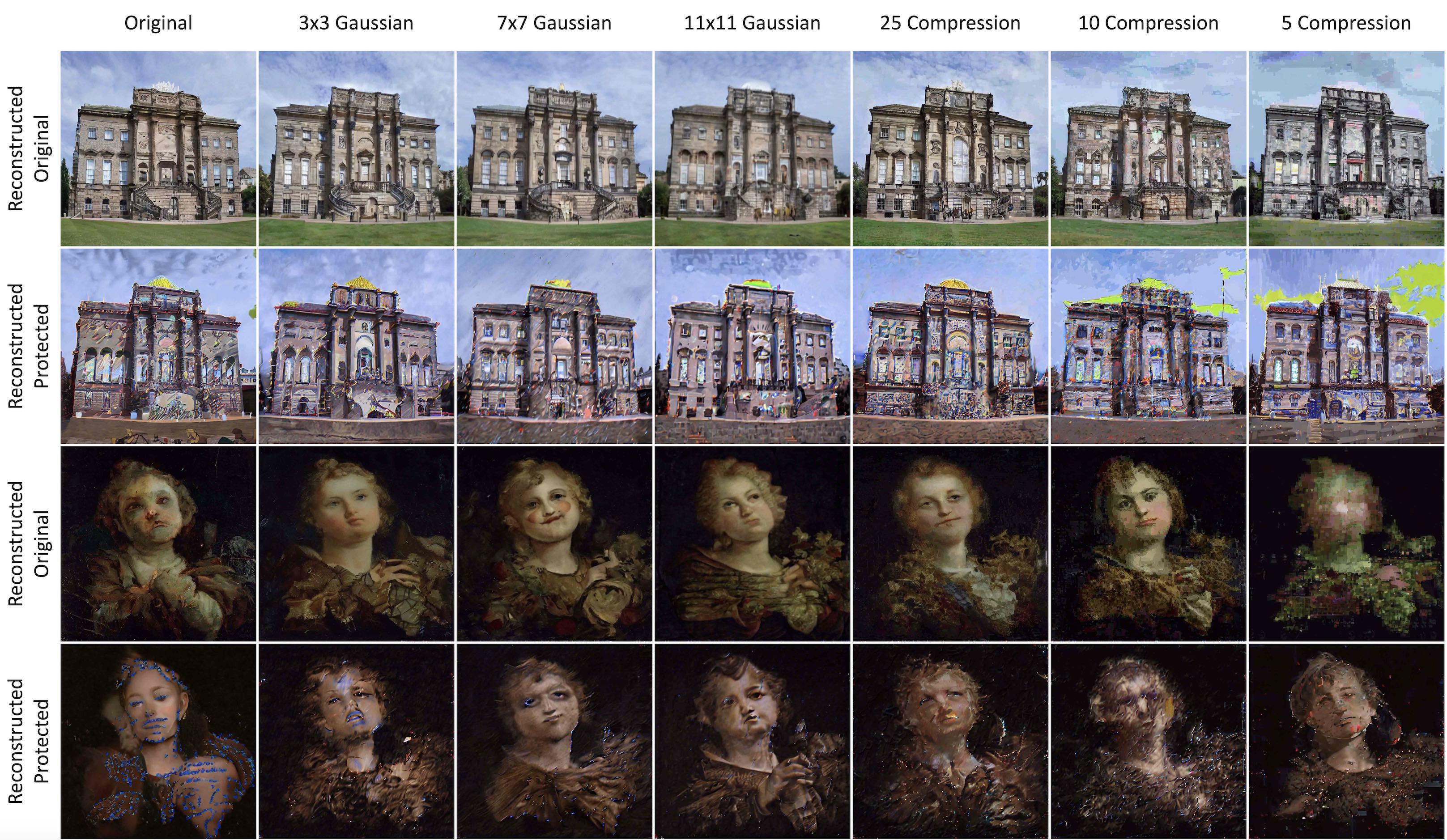} 
\caption{Analyzing the impact of blurring and compressing input and protected images, large blur kernels (11x11) improve diffusion output (contradicts protection) whereas high compression (5) helps confuse diffusion (supports protection).}
\label{fig:post}
\end{figure*}

\begin{table}[h]
\begin{tabular}{ccllll}
\multicolumn{1}{l}{}                                                                          & \multicolumn{1}{l}{}                  & PSNR  & RMSE & SSIM & FID    \\ \cline{2-6} 
\multicolumn{1}{c|}{\multirow{2}{*}{\begin{tabular}[c]{@{}c@{}}Wiki \\ Art\end{tabular}}}     & \multicolumn{1}{c|}{(1)}       & 25.98 & 7.90 & 0.87 & 123.52 \\
\multicolumn{1}{c|}{}                                                                         & \multicolumn{1}{c|}{(2)} & 14.97 & 9.66 & 0.26 & 158.08 \\ \cline{2-6} \hline 
\multicolumn{1}{c|}{\multirow{2}{*}{\begin{tabular}[c]{@{}c@{}}Historic \\ Art\end{tabular}}} & \multicolumn{1}{c|}{(1)}       & 28.15 & 6.36 & 0.88 & 92.83  \\
\multicolumn{1}{c|}{}                                                                         & \multicolumn{1}{c|}{(2)} & 16.24 & 9.42 & 0.32 & 163.80 \\ \hline
\multicolumn{1}{c|}{\multirow{2}{*}{\begin{tabular}[c]{@{}c@{}}Art \\ 201\end{tabular}}}      & \multicolumn{1}{c|}{(1)}       & 24.83 & 7.79 & 0.80 & 209.06 \\
\multicolumn{1}{c|}{}                                                                         & \multicolumn{1}{c|}{(2)} & 15.73 & 9.68 & 0.29 & 241.43 \\
\hline
\multicolumn{1}{c|}{\multirow{2}{*}{\begin{tabular}[c]{@{}c@{}}Face \\ Forensics\end{tabular}}}      & \multicolumn{1}{c|}{(1)}       & 35.06 & 3.82 & 0.95 & 75.15 \\
\multicolumn{1}{c|}{}                                                                         & \multicolumn{1}{c|}{(2)} & 22.40 & 8.81 & 0.73 & 106.96
\end{tabular}
\caption{Quantitative similarities of (1) input and protected images and (2) diffusion outputs of (1), over four datasets.}
\label{tab:all}
\end{table}

\subsection{Evaluations}

Aligned with our assumptions, we want to validate that (1) input and protected images are similar enough and (2) diffusion output of the protected image has low quality. We visualize the success of MAMC in Fig. 1 and document quantitative evaluations in Tab.~\ref{tab:all} in terms of the average PSNR, RMSE, SSIM, and FID for (1) and (2) above; over three aforementioned datasets. For (1), we aim to have ``better'' scores. High PSNR means protected image is not noisy, low RMSE means MAMC reconstruction quality is good, and high SSIM means MAMC preserves the image structurally. On the other hand, all of these scores significantly getting worse for (2) means that diffusion outputs are very different. Especially comparing FID scores indicate that the diffusion output of protected images are indeed adversarial for any model with no representative power.

\subsubsection{Cross Dataset Evaluation.} Following our assumptions, MAMC should support any image, independent of training sets. To support this generalization, we perform cross-dataset evaluations in Tab.~\ref{tab:quan} between three datasets.

\begin{table}[h]
\begin{tabular}{llllllll}
\multicolumn{2}{c}{Training}                                                                                         & \multicolumn{2}{c}{Wiki Art}                         & \multicolumn{2}{c}{His. Art}                         & \multicolumn{2}{c}{Art 201}                       \\ \hline
\multicolumn{2}{c|}{Testing}                                                                                         & \multicolumn{1}{c}{(1)} & \multicolumn{1}{c}{(2)}   & \multicolumn{1}{c}{(1)} & \multicolumn{1}{c}{(2)}   & \multicolumn{1}{c}{(1)} & \multicolumn{1}{c}{(2)} \\ \hline 
\multicolumn{1}{c|}{\multirow{3}{*}{\begin{tabular}[c]{@{}l@{}}Wiki\\ Art\end{tabular}}} & \multicolumn{1}{l|}{PSNR} & 25.                   & \multicolumn{1}{l|}{14.9} & 28.2                   & \multicolumn{1}{l|}{16.7} & 27.5                   & 15.1                   \\
\multicolumn{1}{c|}{}                                                                    & \multicolumn{1}{l|}{RMSE} & 7.90                    & \multicolumn{1}{l|}{9.66}  & 6.41                    & \multicolumn{1}{l|}{9.48}  & 7.17                    & 9.66                    \\
\multicolumn{1}{c|}{}                                                                    & \multicolumn{1}{l|}{SSIM} & 0.87                    & \multicolumn{1}{l|}{0.26}  & 0.86                    & \multicolumn{1}{l|}{0.33}  & 0.89                    & 0.27                    \\ \hline
\multicolumn{1}{l|}{\multirow{3}{*}{\begin{tabular}[c]{@{}l@{}}His.\\ Art\end{tabular}}} & \multicolumn{1}{l|}{PSNR} & 28.2                   & \multicolumn{1}{l|}{16.7} & 28.1                   & \multicolumn{1}{l|}{16.2} & 28.2                   & 16.7                   \\
\multicolumn{1}{l|}{}                                                                    & \multicolumn{1}{l|}{RMSE} & 6.41                    & \multicolumn{1}{l|}{9.48}  & 6.36                    & \multicolumn{1}{l|}{9.42}  & 6.41                    & 9.48                    \\
\multicolumn{1}{l|}{}                                                                    & \multicolumn{1}{l|}{SSIM} & 0.86                    & \multicolumn{1}{l|}{0.33}  & 0.88                    & \multicolumn{1}{l|}{0.32}  & 0.86                    & 0.33                    \\ \hline
\multicolumn{1}{l|}{\multirow{3}{*}{\begin{tabular}[c]{@{}l@{}}Art\\ 201\end{tabular}}}  & \multicolumn{1}{l|}{PSNR} & 27.5                   & \multicolumn{1}{l|}{15.1} & 28.2                   & \multicolumn{1}{l|}{16.7} & 24.8                   & 15.7                   \\
\multicolumn{1}{l|}{}                                                                    & \multicolumn{1}{l|}{RMSE} & 7.17                    & \multicolumn{1}{l|}{9.66}  & 6.41                    & \multicolumn{1}{l|}{9.48}  & 7.79                    & 9.68                    \\
\multicolumn{1}{l|}{}                                                                    & \multicolumn{1}{l|}{SSIM} & 0.89                    & \multicolumn{1}{l|}{0.27}  & 0.86                    & \multicolumn{1}{l|}{0.35}  & 0.80                    & 0.29                   
\end{tabular}
\caption{Cross-dataset evaluations between three datasets measured on (1) input and protected images and (2) diffusion output of input and protected images.}
\label{tab:quan}
\end{table}

\subsection{Robustness}
In order to assess the robustness of our adversarial protection, we apply different levels of Gaussian blur and compression to input and protected images, then visualize the diffusion outputs of post-processed images in Fig.~\ref{fig:post}. Similar to claims of~\cite{photoguard}, compression helps with confusing diffusion process, as observed in the last column. As MAMC aims to preserve the image perceptually, perturbations are most likely inserted in higher frequencies of the image. Thus, large kernel blurs (as in the 4th column) degrade our protection, creating more similar diffusion outputs. Nevertheless, noise and structural degradation do exist.

\subsection{Ablation Studies}\label{sec:abla}
We analyze the contribution of each loss term, the balance of reconstruction hyper-parameters, and the effect of diffusion strength in this section.

\subsubsection{Diffusion Strength.} 
\begin{figure}[h]
\centering
\includegraphics[width=1\linewidth]{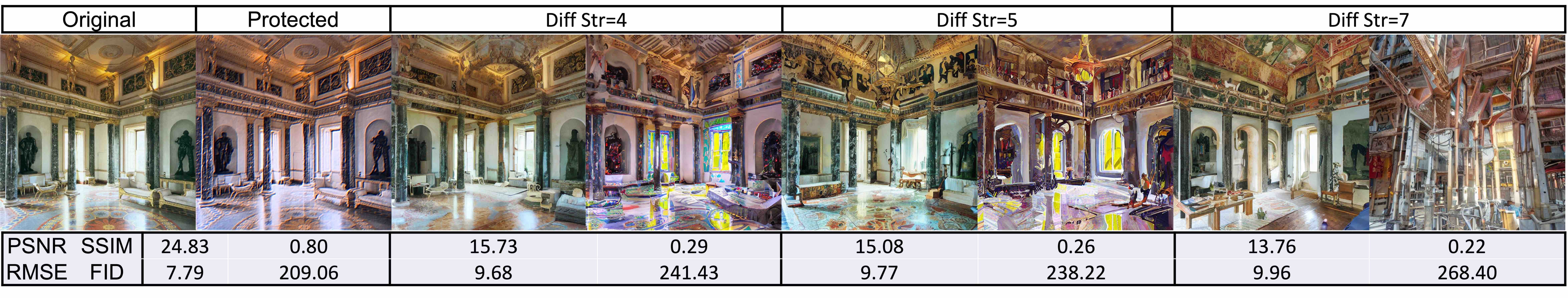} 
\caption{We analyze the impact of diffusion strength qualitatively and quantitatively.}
\label{fig:sd}
\end{figure}

During our experiments, we also analyzed the impact of the diffusion strength parameter. The larger the value is, the less guidance the model incorporates, which means less protection is needed as the imitation pressure of the model is decreasing. We verify this by changing diffusion parameters as depicted in Fig.~\ref{fig:sd}, and calculating the similarity scores of diffusion outputs of protected and input images (as indicated by (2) in previous experiments). We start with the input similarity scores (as indiacted by (1) in previous experiments) and images as reference; then analyze similarity scores of diffusion outputs with $str=\{4,5,7\}$ values. Results confirm our hypothesis visually and quantitatively. We use $str=5$ to balance fidelity and guidance.

\subsubsection{Loss Contributions.} 
\begin{figure}[h]
\centering
\includegraphics[width=1\linewidth]{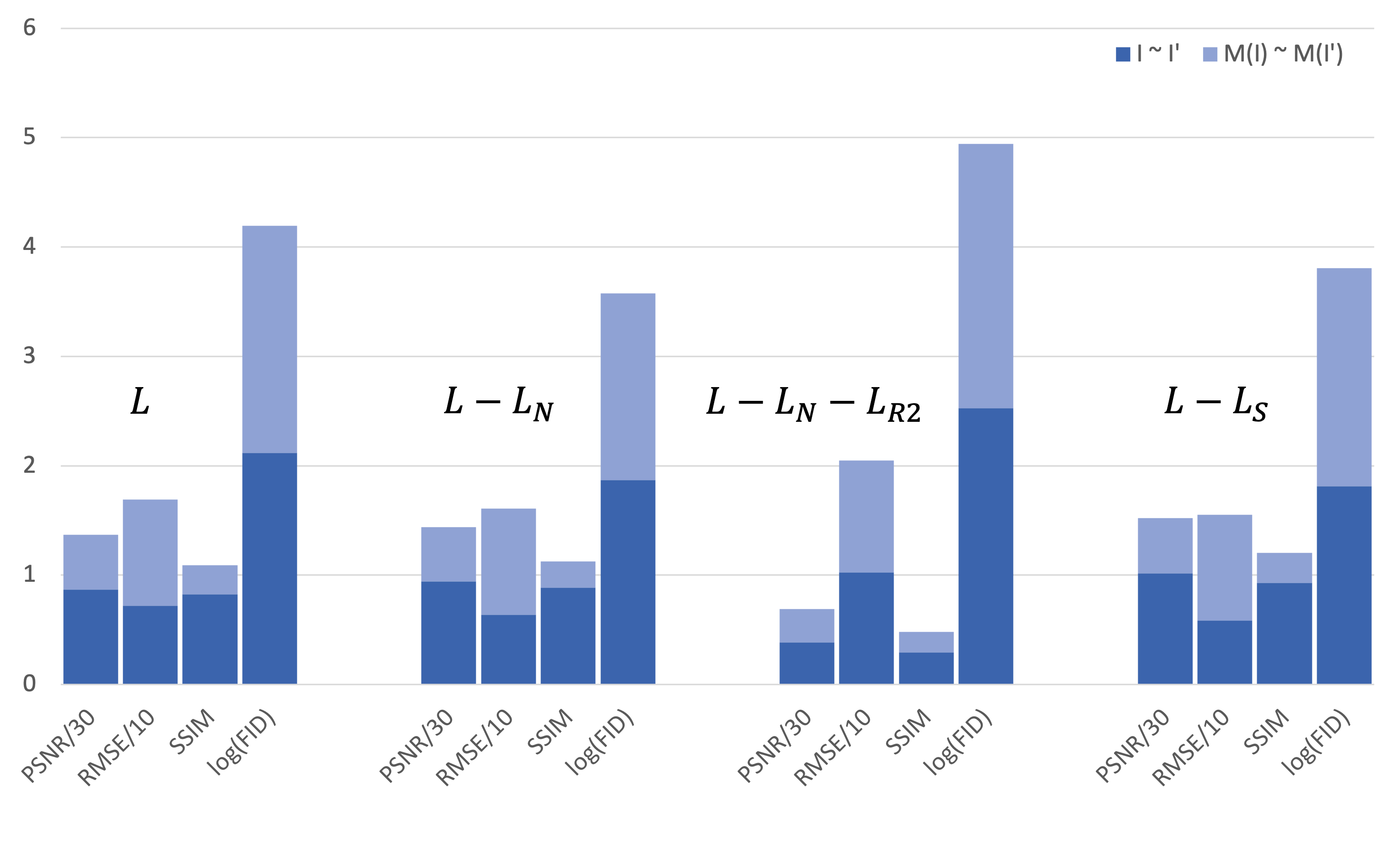} 
\caption{Four experimental loss settings are compared in terms of noise, pixel, structure, and generative spaces.}
\label{fig:loss}
\end{figure}

We design four experiments to understand our loss landscape: overall loss $L$, without noise loss ($L-L_N$), without regularization ($L-L_N-L_{R2}$), and without style loss ($L-L_S$). In Fig.~\ref{fig:loss}, we simplify the graphs by normalizing PSNR by 30, RMSE by 10, and plot FID in log scale. We observe that noise loss restricts the representative power and pixel-wise regularization is indeed needed to preserve structure and content. We do not see significant quantifiable impact of the style loss, however, visual inspection hints its need.

\subsubsection{Reconstruction Weights.} We observe that $\alpha_{R1}$ and $\alpha_{R2}$ maintain an intricate balance, hence we experiment with different values with $\alpha_{R2}=\{0.75, 1, 1.5\}$ in Tab. 3.
\begin{table}[h]\label{tab:rec}
\begin{tabular}{l|ll|ll|ll}
$\alpha_{R2}$   & \multicolumn{2}{c|}{0.75} & \multicolumn{2}{c|}{1.0} & \multicolumn{2}{c}{1.5} \\ \hline
Setting & (1)         & (2)         & (1)        & (2)       & (1)        & (2)        \\ \hline
PSNR    & 22.95       & 14.26       & 28.30      & 14.90     & 29.70      & 15.65      \\
RMSE    & 8.43        & 9.82        & 6.37       & 9.72      & 6.01       & 9.69       \\
SSIM    & 0.79        & 0.24        & 0.89       & 0.24      & 0.90       & 0.29       \\
FID     & 167.6      & 228.7      & 74.35      & 50.86     & 96.50      & 184.5    
\end{tabular}

\caption{Reconstruction regularization weights are analyzed.}
\end{table}

\subsection{Implementation Details}  
\label{sec:curve}
\begin{figure}[h]
\centering
\includegraphics[width=1\linewidth]{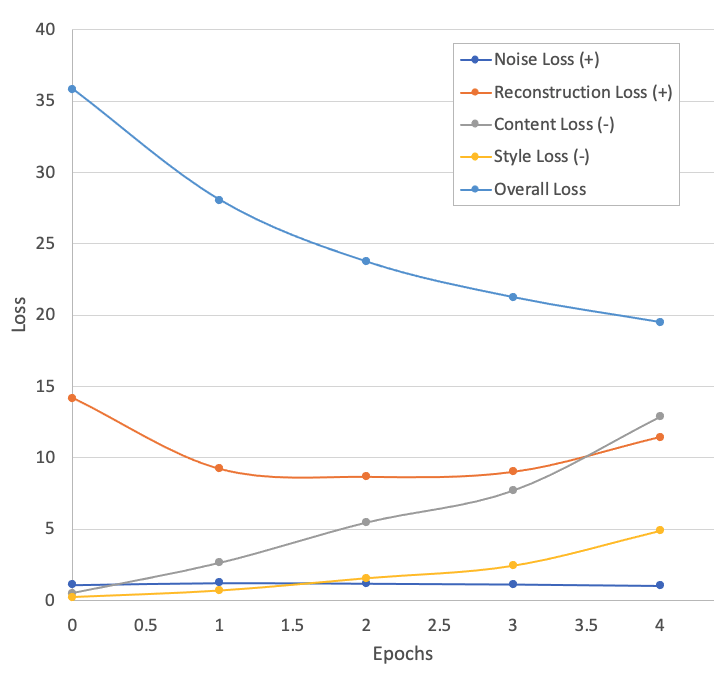} 
\caption{Plotting loss terms' contributions to the overall loss among consecutive epochs.}
\label{fig:lossplot}
\end{figure}

We set U-Net's learning rate to 0.001. Similar to other normalizations leading to effective training, we take each art piece and adjust and distribute the value of each coordinate's RGB value to be between 0 and 1. As mentioned previously we utilize 1000 and 5000 subsets of WikiArt and Historic Art datasets, where each is partitioned into two distinct subsets adhering 70\% to training and 30\% to testing. We process this data through only a few epochs to enhance the model's ability of developing adversarial representation, as demonstrated in Fig.~\ref{fig:lossplot}. Although our loss landscape looks non-convex with negative and positive terms, we observe that it decreases. Note that some terms have negative effects to the overall loss. Experiments are done on NVIDIA GTX 1080 TI with experiments taking approximately an hour for the 1000 subset and 6 hours for the 5000 subset.

\section{Conclusion} 
We present ``My Art My Choice'', an adversarial protection model to prevent online images from being exploited by diffusion models. We believe there is indeed a need for online protection of copyrighted material and our cross-domain protector is ideal for interrupting many diffusion-based tasks as we demonstrated, such as personalization, style transfer, and any guided image-to-image translation. We evaluate MAMC on four different datasets, with cross-dataset settings, relating to ablation studies on loss terms and hyper-parameters, and assessing user control. 
As generative AI services are only in their early days, we belive proactive protection services based on MAMC will also be valuable in future. That being said, our next research direction is to apply MAMC to other modalities and domains that are in need of provenance protection. Meanwhile, we also plan to invest in red teaming our adversarial protection approach to make it stronger against unseen and unknown adversaries. 

\bibliography{aaai24}

 \begin{figure*}[ht]
 \centering
 \includegraphics[width=1\linewidth]{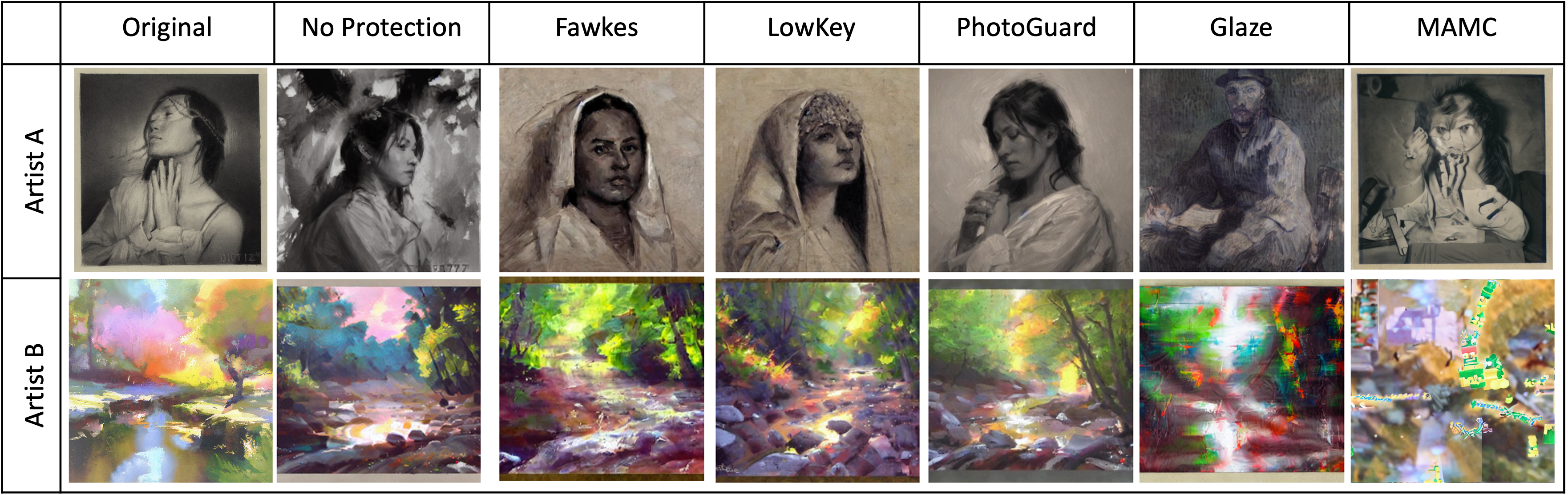} 
 \caption{MAMC compared visually against image cloaking approaches.}
 \label{fig:comp}
 \end{figure*}

\newpage
\appendix
\makeatletter
\renewcommand{\@seccntformat}[1]{Appendix \csname the#1\endcsname\quad}
\makeatother
\renewcommand{\thesection}{\Alph{section}}

\section{Comparison}\label{app:A}
We compare MAMC to image cloaking approaches such as ~\cite{fawkes,lowkey,photoguard,glaze} on the 1K subset of Wiki Art~\cite{wikiart} dataset. As we do not have access to the same experimental setup of ~\cite{glaze} to report CLIP-based genre shift scores, we demonstrate the results visually in Figure~S\ref{fig:comp}, with the same sample used in~\cite{glaze}. MAMC seems to push the protected image to cause a significantly ``bad'' diffusion output. Note that, images created by other approaches are based on text-guidance like ``A girl in the style of Karla Ortiz, black and white''. In contrast, our target is to eliminate the image being used for guidance, so we expect the diffusion output to be as distorted as possible, which is reflected on the downstream editing/generation tasks. Furthermore, if the bad actors using diffusion models are not familiar with the artist's style, they may still distribute outputs created from these cloaked images as in the artists' style without recognizing the difference, which is also damaging. For MAMC, the output is so distorted that it is obviously not a useful image, in the style of nobody.

\section{Use Case: Inpainting}\label{app:B}
\begin{figure}[h!]
\centering
\includegraphics[width=1\linewidth]{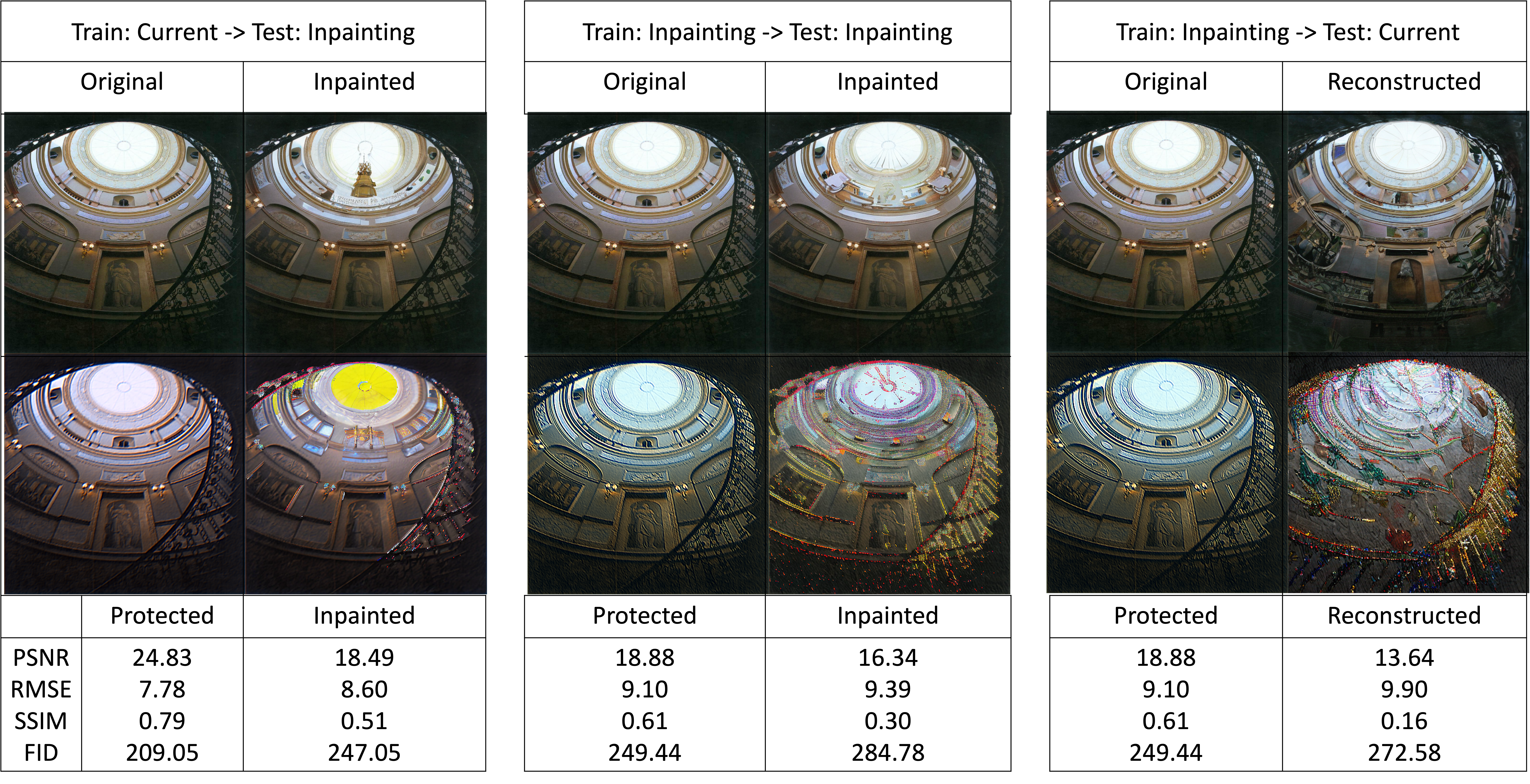} 
\caption{(Left) Previous model tested for inpainting, (middle) model retrained and tested for inpainting, (right) retrained model tested for reconstruction.}
\label{fig:in}
\end{figure}

Another general use for diffusion models is scene manipulation, by copying and pasting objects, or simply by infilling masked areas. In order to verify our claim that MAMC provides adequate protection in a task-agnostic manner, we evaluat our approach on three scenarios: (1) testing the pretrained model mentioned in the main text on inpainting task, (2) training and testing our generator on inpainting task, and (3) testing this new model on the old task of reconstruction. Figure~S\ref{fig:in} demonstrates samples from these three scenarios, where each box contains input image, protected image, diffusion output of the input image, and diffusion output of the protected image. For the first two scenarios, diffusion is used for inpainting a mask of 120x66 pixels in the middle of upper half of the image
. We document same metrics following the main paper.

We observe that the current model is able to disable inpainting, almost as well as disabling content replication. In addition, when the model is specifically trained for inpainting task, protected image metrics get worse. We hypothesize that when a part of the input image is missing, pixel-wise regularization of the reconstruction loss dominates, creating grainy ``pixel perfect'' images. We also remark that the model trained on inpainting and used for reconstruction creates even worse diffusion outputs, yielding yet another possibility for the user to increase protection.

\end{document}